\documentclass[10pt,twocolumn,letterpaper]{article}

\usepackage{cvpr}              

\usepackage{times}
\usepackage{epsfig}
\usepackage{graphicx}
\usepackage{amsmath}
\usepackage{amssymb}

\definecolor{cvprblue}{rgb}{0.21,0.49,0.74}
\usepackage[pagebackref,breaklinks,colorlinks,allcolors=cvprblue]{hyperref}

\def\paperID{17} 
\def\confName{CVPR}
\def\confYear{2025}

\definecolor{grayhighlight}{RGB}{213,229,255}

\newsavebox\CBox

\begin{document}

\title{4K Image Super-Resolution on Mobile NPUs: Mobile AI 2026 Challenge Report}

\author{
Andrey Ignatov \and Georgy Perevozchikov \and Radu Timofte 
\and Fubin Ren \and Shihao Wang
\and Liwen Zhang \and Zhe Xu \and Zhipei Lei \and Minjian Zhang \and Kailing Tang \and Dingyong Gou \and Yanlin Wu \and Cong Li \and Xiaohui Cui
\and Kihwan Yoon \and Ganzorig Gankhuyag
\and Heo MyeongCheol \and Yonghyun Park \and Juho Jung \and Sungjei Kim
\and Phuong-Nam Nguyen-Pham \and Nam-Tien Le \and Tinh-Anh Nguyen-Nhu
\and Matin Fazel \and Abdelhak Bentaleb
\and Xinrui Li \and Xusheng He \and Jianlong Wu
\and Mohsin Ikram \and Dmitry Ignatov
\and Hao Chen \and Long Sun \and Jiangxin Dong \and Jinshan Pan \and Jinhui Tang
\and Siyue Teng \and Yuxuan Jiang \and Fan Zhang \and David Bull
\and Zhe Xiao \and Jun Xiao \and Zhi-Song Liu \and Jiayi Pei \and Songtao Wu \and Hong Shang \and Kuanhong Xu
}

\maketitle

\begin{abstract}
	
Image super-resolution is widely used on mobile devices for upscaling various types of image data. This creates a need for efficient deep learning-based models that are not only performant but can also be deployed on neural processing units (NPUs) present in the majority of mobile and IoT chipsets. In this Mobile AI challenge, we address this problem and propose the participants to design fast solutions that can upsample images from Full HD to 4K resolution while running on popular mobile NPUs from Qualcomm and MediaTek. For this, the participants were provided with the DIV2K dataset and trained models to do an efficient 3X image upscaling. The competition had two tracks: in the first track, the models had to be fully quantized (INT8), and the runtime of all models was evaluated on the Qualcomm Snapdragon 8 Elite Gen 5 NPU. In the second track, the participants were developing floating-point solutions that were evaluated on the MediaTek Dimensity 9500 NPU. Top solutions proposed in this challenge are able to reconstruct 4K images under 5 milliseconds while reaching a PSNR score of 30 dB. A comprehensive description of the models developed in the challenge is provided in this paper.

\end{abstract}
{\let\thefootnote\relax\footnotetext{%
\hspace{-5mm}$^*$Andrey Ignatov, Georgy Perevozchikov and Radu Timofte are the organizers of the Mobile AI 2026 challenge \textit{(andrey@vision.ee.ethz.ch, georgii.perevozchikov@uni-wuerzburg.de, radu.timofte@uni-wuerzburg. de)}. The other authors participated in the challenge. \\ Appendix \ref{sec:apd:team} contains the authors' team names and affiliations.\\ Mobile AI 2026 Workshop website: \\ \url{https://ai-benchmark.com/workshops/mai/2026/}
}}

\section{Introduction}

\begin{figure*}[t!]
\centering
\setlength{\tabcolsep}{1pt}
\resizebox{\linewidth}{!}
{
\includegraphics[width=0.5\linewidth]{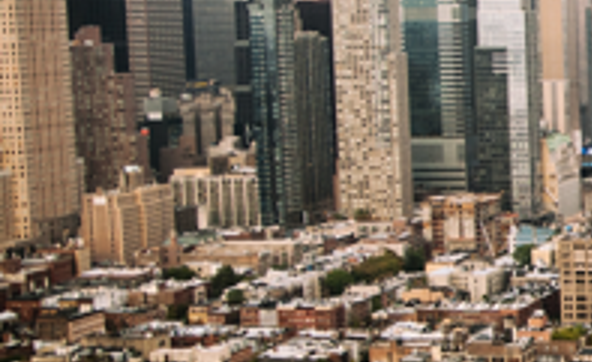} \hspace{2mm}
\includegraphics[width=0.5\linewidth]{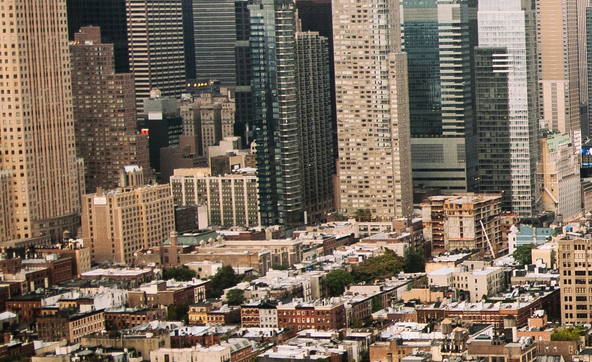}
}
\vspace{0cm}
\caption{Sample crops from a 3X bicubically upscaled image and the target DIV2K~\cite{agustsson2017ntire} photo.}
\label{fig:example_photos}
\end{figure*}

Single-image super-resolution is a core computer vision task that aims to generate a high-resolution image from its downsampled version by restoring fine details and high-frequency information. Unlike the standard image interpolation techniques~\cite{irani1991improving,freeman2002example,park2003super,timofte2013anchored,timofte2014a+,yang2013fast,yang2008image,yang2010image,huang2015single,timofte2016seven}, deep learning-based approaches are able to achieve significantly superior results in terms of both reconstruction fidelity and perceptual quality. In recent years, this task has gained significant attention due to its direct applications in telephoto image processing and AI-based digital zoom in smartphone cameras, enhancement of low-resolution media, frame interpolation in video games, or upscaling of images and videos to match the resolution of display panels. A wide range of deep learning-based methods have been proposed in the past to achieve high-fidelity image reconstruction~\cite{dong2015image,dong2014learning,kim2016accurate,lim2017enhanced,timofte2017ntire,timofte2018ntire,cai2019ntire,
lugmayr2020ntire,zhang2020ntire,ignatov2018pirm,liang2021swinir,chen2022cross,zhang2022efficient,chen2023recursive,chen2023activating,ren2025tenth}. However, a primary limitation of these approaches is that they are typically focused on optimizing the restoration quality, while neglecting the computational efficiency required for mobile platforms, where resource constraints are a critical consideration. In this challenge, we directly address this problem and propose the participants to design efficient super-resolution models that can run on recent smartphone neural processing units (NPUs) from Qualcomm and MediaTek mobile chipsets.

Designing deep learning models that run efficiently on mobile devices~--- especially across different AI accelerators like NPUs~--- is a complex challenge. It demands careful analysis of hardware constraints such as limited operator support, restricted memory capacity and bandwidth, reduced accuracy for quantized models and specific driver behavior of NPUs. Even when certain layers or operations are technically supported, they may be executed inefficiently due to constraints related to NPU hardware architecture or its driver implementation, and thus create performance bottlenecks during inference. A striking example here would be the Google Tensor NPU present in all recent Google Pixel smartphones that, despite having quite high theoretical peak performance, is unable to efficiently accelerate the vast majority of even simplistic CNN-based models~\cite{ignatov2025MAISR}. As a result, achieving low latency and energy efficiency often requires partial or complete redesigns of model architectures. For a more detailed discussion of these challenges and the performance of modern mobile AI accelerators we refer to~\cite{ignatov2019ai,ignatov2018ai}. Overall, all these limitations require various adaptations of the standard deep learning models for their efficient deployment on mobile NPUs, including conventional optimizations like 16-bit / 8-bit~\cite{chiang2020deploying,jain2019trained,jacob2018quantization,yang2019quantization,li2021mqbench,nagel2021white,bondarenko2021understanding,siddegowda2022neural,tan2024mobilequant,lin2025awq} and low-bit~\cite{cai2020zeroq,uhlich2019mixed,banner2019post,ignatov2020controlling,liu2018bi,zhuo2022empirical,bhalgat2020lsq+,kim2023memory,dettmers2023case,bondarenko2024low,li2025memory} quantization, network pruning and compression~\cite{chiang2020deploying,li2019learning,
liu2019metapruning,obukhov2020t,kim2021pqk,hawks2021ps,zheng2022model,xu2023sparsebytenn,kuzmin2023pruning,wang2024towards}, NPU-specific adaptations, platform-aware neural architecture search~\cite{howard2019searching,tan2019mnasnet,wu2019fbnet,
wan2020fbnetv2,mills2021profiling,chu2021discovering,zhan2021achieving,wu2022compiler,chitty2022neural,yoshihama2023hardware}, \etc.

In this Mobile AI Challenge, the participants were proposed to develop efficient deep learning super-resolution models that can upscale Full HD images to 4K resolution directly on smartphone NPUs. The DIV2K~\cite{agustsson2017ntire} dataset containing diverse RGB images was provided for network training, a  downscaling factor of 3 was considered in this competition. The competition had two tracks: the first one, where the standard floating-point (FP16) models were considered, and the second one, where all models had to be additionally INT8 quantized (both weights and activations) to ensure high efficiency and compatibility of the designed solutions with low-power edge NPUs. The runtime of the models was evaluated on the Dimensity 9500 NPU and the Snapdragon 8 Elite Gen 5 NPU in the 1st and 2nd tracks, respectively. Each submission was evaluated based on both image fidelity and runtime efficiency, thus balancing reconstruction quality with model efficiency. All solutions developed during this challenge are fully compatible with the LiteRT~\cite{LiteRT2025} (former TensorFlow Lite) framework, enabling deployment on a variety of Linux and Android-based IoT platforms, smartphones, and edge devices.

\smallskip


This challenge is a part of the \textit{MAI 2026 Workshop and Challenges} consisting of the following competitions:



\begin{itemize}
\item 4K Image Super-Resolution on Mobile NPUs
\item Real-Time Video Super-Resolution on Smartphone GPUs~\cite{ignatov2025srvideo}
\item Efficient Image Denoising on Smartphone GPUs~\cite{ignatov20254denoising}
\item RGB Photo Enhancement on Smartphone GPUs~\cite{ignatov2025rgb}
\item Learned Smartphone ISP on Mobile GPUs~\cite{ignatov2025isp}
\item Bokeh Effect Rendering on Smartphone GPUs~\cite{ignatov2025bokeh}
\item Efficient LLMs for On-Device Deployment
\item Adapting Stable Diffusion for Efficient On-Device Inference
\end{itemize}

\normalsize


\noindent The results and solutions obtained in the previous \textit{Mobile AI Challenges} are described in the following papers:


\begin{itemize}
\item Image Super-Resolution on Mobile NPUs~\cite{ignatov2021real,ignatov2022efficient,ignatov2025MAISR}
\item Learned ISP on Mobile NPUs and GPUs~\cite{ignatov2021learned,ignatov2022learned,ignatov2025MAIRAW}
\item Video Super-Resolution on Mobile GPUs~\cite{romero2021real,ignatov2022power}
\item Image Denoising on Mobile GPUs~\cite{ignatov2021fastDenoising}
\item Bokeh Effect Rendering on Mobile GPUs~\cite{ignatov2022realistic}
\item Camera Scene Detection on Smartphones~\cite{ignatov2021fastSceneDetection,pouget2021fast}
\item RGB Photo Enhancement on Mobile GPUs~\cite{ignatov2025MAIRGB}
\item Depth Estimation on Mobile Devices~\cite{ignatov2021fastDepth,ignatov2022efficient}
\end{itemize}

\normalsize

\begin{figure*}[t!]
\centering
\setlength{\tabcolsep}{1pt}
\resizebox{0.96\linewidth}{!}
{
\includegraphics[width=1.0\linewidth]{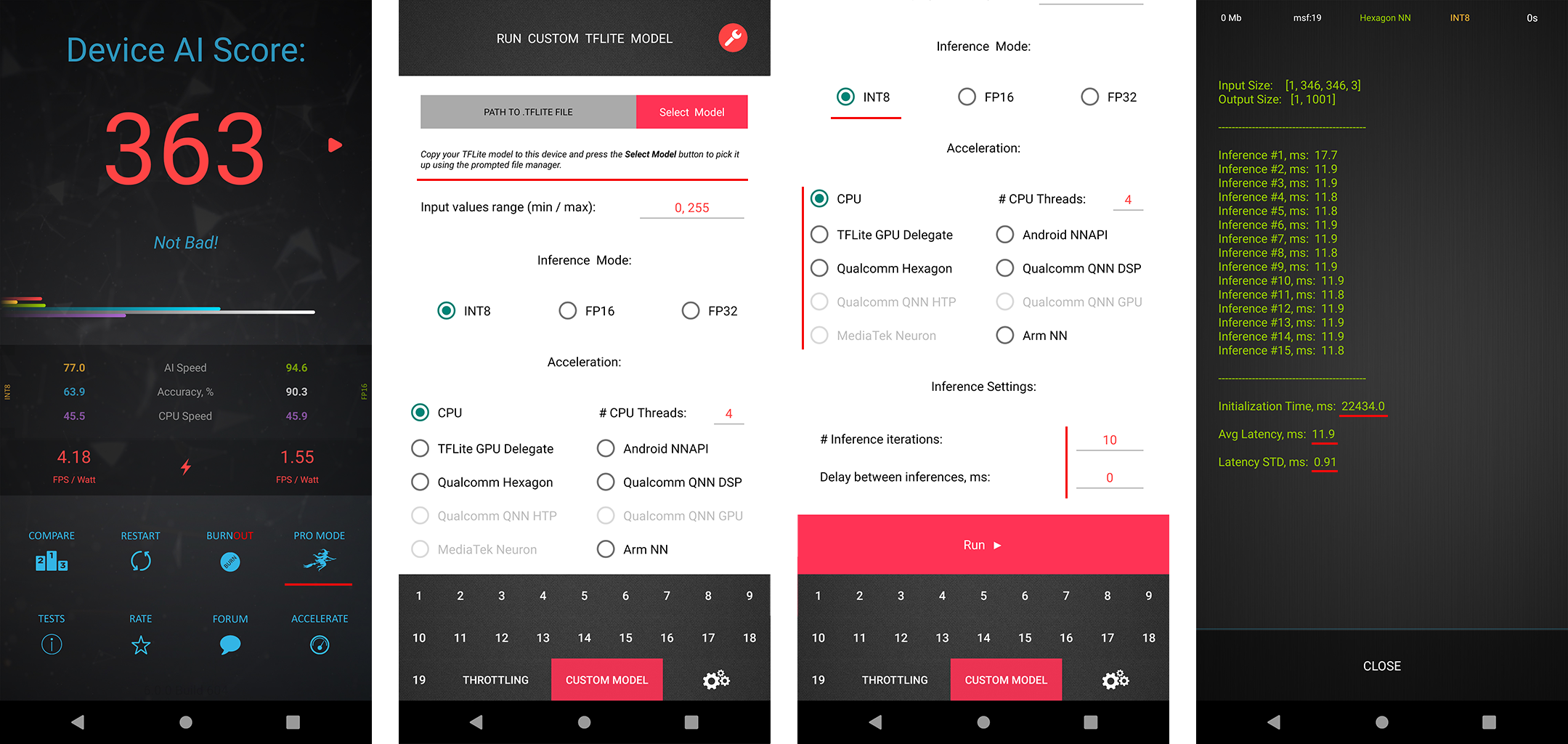}
}
\vspace{0.2cm}
\caption{Loading and running custom TensorFlow Lite models with AI Benchmark application. The currently supported acceleration options include Qualcomm QNN, Hexagon NN, MediaTek Neuron, TFLite GPU, Android NNAPI and ARM NN delegates as well as CPU inference (LiteRT / XNNPACK backends). The latest app version can be downloaded at \url{https://ai-benchmark.com/download}}
\label{fig:ai_benchmark_custom}
\end{figure*}

\section{Challenge}

To design an efficient and practical solution for mobile-related tasks, the following key components are essential:


\begin{enumerate}
\item A high-quality and large-scale dataset that can be used to train and evaluate the solution;
\item An efficient way to check the runtime and debug the model locally without any constraints;
\end{enumerate}

This challenge addresses the above issues. Real training data, tools, and runtime evaluation options provided to the challenge participants are described in the next sections.

\subsection{Dataset}

In this challenge, the participants were proposed to work with the popular DIV2K~\cite{agustsson2017ntire} dataset. It consists of 1000 diverse 2K resolution RGB images: 800 are used for training, 100 for validation and 100  for testing purposes. The images are of high quality both aesthetically and in terms of small amounts of noise and other corruptions (like blur and color shifts). All images were manually collected and have 2K pixels on at least one of the axes (vertical or horizontal). DIV2K covers a large diversity of contents, from people, handmade objects and environments (cities), to  flora and fauna and natural sceneries, including underwater. An example set of images is demonstrated in Fig.~\ref{fig:example_photos}.

\subsection{Local Runtime Evaluation}

When developing AI solutions for mobile devices, it is vital to be able to test the designed models and debug all emerging issues locally on available devices. For this, the participants were provided with the \textit{AI Benchmark} application~\cite{ignatov2018ai,ignatov2019ai} that allows to load any custom TensorFlow Lite model and run it on any Android device with all supported acceleration options. This tool contains the latest versions of the \textit{Qualcomm QNN, MediaTek Neuron, Hexagon NN, TFLite GPU, Android NNAPI, Samsung ENN, Amlogic AML} and \textit{Arm NN} delegates, therefore supporting all current mobile platforms and providing the users with the ability to execute neural networks on smartphone NPUs, APUs, DSPs, GPUs and CPUs.

\smallskip

To load and run a custom TensorFlow Lite (LiteRT) model, one needs to follow the next steps:

\begin{enumerate}
\setlength\itemsep{0mm}
\item Download AI Benchmark from the official website\footnote{\url{https://ai-benchmark.com/download}} or from Google Play\footnote{\url{https://play.google.com/store/apps/details?id=org.benchmark.demo}} and run its standard tests.
\item After the end of the tests, enter the \textit{PRO Mode} and select the \textit{Custom Model} tab there.
\item Put the exported TFLite (LiteRT) model into the \textit{Download} folder of the device.
\item Select mode type \textit{(INT8, FP16, or FP32)}, the desired acceleration / inference options and run the model.
\end{enumerate}

\noindent These steps are also illustrated in Fig.~\ref{fig:ai_benchmark_custom}. The instructions on how to generate TFLite model from PyTorch, TensorFlow and Keras pre-trained models were provided in the MAI Workshop GitHub repository~\footnote{\url{https://github.com/aiff22/MAI-2025-Workshop}}.

\subsection{Runtime Evaluation on the Target Platform}

In this challenge, we use the two flagship mobile SoCs as our target evaluation platforms: \textit{Qualcomm Snapdragon 8 Elite Gen 5} and \textit{MediaTek Dimensity 9500} chipsets, both featuring dedicated neural processing units. The considered NPUs demonstrates the highest AI Benchmark scores at the beginning of year 2026~\cite{AIBenchmark202603}, and can be found in numerous flagship smartphones. The models were parsed and accelerated using the Qualcomm QNN and MediaTek Neuron TFLite delegates on the Snapdragon 8 Elite Gen 5 and the Dimensity 9500 SoC, respectively. All final solutions were tested using the aforementioned AI Benchmark application.

\begin{table*}[t!]
\centering
\resizebox{\linewidth}{!}
{
\begin{tabular}{l|c|cc|ccc|ccc|c}
\hline
Team \, & \, Author \, & \, Framework \, & \, Model Size, \, & \, PSNR$\uparrow$ \, & \, SSIM$\uparrow$ \, & \, $\Delta$ PSNR Drop, \, & \multicolumn{2}{c}{\, Runtime, ms $\downarrow$ \,} & Speed-Up & \, Final Score \\
& & & \small{KB} & \multicolumn{2}{c}{\small{INT8 Model}} & \small{FP32 $\to$ INT8} & \, \small{CPU} \, & \, \small{NPU} \, & & \\
\hline
\hline
			AntYSP & Re4Zen1 & \, PyTorch / TensorFlow \, & 54.9 & 29.96 & 0.8729 & 0.08 & 338 & 4.33 & 78.1 & 21.8 \\
AntSR$^*$ & zzyAnt & PyTorch / TensorFlow & 65.3 & 29.98 & 0.8731 & 0.1 & 404 & 4.52 & 89.4 & 21.5 \\
z6 & rlghksdbs & TensorFlow & 67.4 & 29.93 & 0.8699 & n.a. & 330 & 5.49 & 60.1 & 16.5 \\
koreatechV2 & HeoMyeongCheol & TensorFlow & 77.1 & 29.06 & 0.8499 & n.a. & 464 & 4.35 & 106.7 & 6.2 \\
AIO\_MAI & namdeptrai2005 & PyTorch & 96.1 & 29.79 & 0.8634 & 0.19 & 610 & 41.1 & 14.8 & 1.8 \\
IN2GM & matin\_fazel & TensorFlow & 64.3 & 29.85 & 0.8701 & 0.03 & 526 & 46.5 & 11.3 & 1.7 \\
LoongSR & felix\_lee & PyTorch & 708 & 29.76 & 0.8586 & 0.67 & 5729 & 5729 & 1 & 0.01 \\
Godzilla & ikrammohsin & PyTorch & 16253 & 30.39 & 0.8831 & n.a. & 215160 & 215160 & n.a. & 0.0008 \\
\hline
A+ regression~\cite{timofte2014a+} & Baseline &  &  & 29.32 & 0.8520 & - & - & - & -  & - \\
Bicubic Upscaling & Baseline &  &  & 28.26 & 0.8277 & - & - & - & - & - \\
\hline
\end{tabular}
}
\vspace{2.6mm}
\caption{\small{Mobile AI 2026 Quantized 4K Image Super-Resolution challenge results and final rankings. During the runtime measurements, the models were performing image upscaling from 720$\times$1280 to 2160$\times$3840 pixels on the Snapdragon 8 Elite Gen 5 SoC. $\Delta$ PSNR values correspond to accuracy loss measured in comparison to the original floating-point network. Team \textit{AntYSP} is the challenge winner.}}
\label{tab:results}
\end{table*}

\subsection{Challenge Phases}

The challenge consisted of the following phases:

\vspace{-0.8mm}
\begin{enumerate}
\item[I.] \textit{Development:} the participants get access to the data and AI Benchmark app, and are able to train the models and evaluate their runtime locally;
\item[II.] \textit{Validation:} the participants can upload their models to the remote server to check the fidelity scores on the validation dataset, and to compare their results on the validation leaderboard;
\item[III.] \textit{Testing:} the participants submit their final results, codes, TensorFlow Lite models, and factsheets.
\end{enumerate}
\vspace{-0.8mm}

\subsection{Scoring System}

All solutions were evaluated using the following metrics:

\begin{itemize}
\setlength\itemsep{-0.2mm}
\item Peak Signal-to-Noise Ratio (PSNR) measuring fidelity score,
\item Structural Similarity Index Measure (SSIM), a proxy for perceptual score,
\item The runtime on the target Snapdragon 8 Elite Gen 5 or Dimensity 9500 NPU.
\end{itemize}

The score of each final submission was evaluated based on the next formula ($C$ is a constant normalization factor):

\smallskip
\begin{equation*}
\text{Final Score} \,=\, \frac{2^{2 \cdot \text{PSNR}}}{C \cdot \text{runtime}},
\end{equation*}
\smallskip

During the final challenge phase, the participants did not have access to the test dataset. Instead, they had to submit their final TensorFlow Lite models that were subsequently used by the challenge organizers to check both the runtime and the fidelity results of each submission under identical conditions. This approach solved all the issues related to model overfitting, reproducibility of the results, and consistency of the obtained runtime / accuracy values.

\section{Challenge Results}

\begin{table*}[t!]
	\centering
	\resizebox{\linewidth}{!}
	{
		\begin{tabular}{l|c|cc|cc|ccc|c}
			\hline
			Team \, & \, Author \, & \, Framework \, & \, Model Size, \, & \, PSNR$\uparrow$ \, & \, SSIM$\uparrow$ \, & \multicolumn{2}{c}{\, Runtime, ms $\downarrow$ \,} & Speed-Up & \, Final Score \\
			& & & \small{KB} & \multicolumn{2}{c}{\small{FP16 Model}} & \, \small{CPU} \, & \, \small{NPU} \, & & \\
			\hline
			\hline
			ZTE\_cv & try\_one\_try & \, PyTorch / TensorFlow \, & 1689 & 30.91 & 0.8921 & 5507 & 110 & 50.1 & 32.1 \\
			AntYSP & re4zen & PyTorch & 2714 & 30.69 & 0.8886 & 7457 & 105 & 71 & 24.8 \\
			Sunrise2046 & chanson & PyTorch & 1452 & 30.61 & 0.8863 & 5541 & 594$^*$ & 9.3 & 3.9 \\
			OrbSR & xeno73 & PyTorch & 3482 & 30.76 & 0.889 & 21537 & 1666$^*$ & 12.9 & 1.7 \\
			BVIISR & st\_eng & PyTorch & 551 & 28.43 & 0.8292 & 2483 & 2106$^*$ & 1.2 & 0.05 \\
			LoongSR & felix\_lee & PyTorch & 2282 & 30.13 & 0.875 & 31906 & 26750$^*$ & 1.2 & 0.04 \\
			Sony-PolyU-LUT & zacharyxiao & PyTorch & 1437 & 28.67 & 0.8448 & 18315 & 12263$^*$ & 1.5 & 0.01 \\
			\hline
			A+ regression~\cite{timofte2014a+} & Baseline &  &  & 29.32 & 0.8520 & - & - & - & - \\
			Bicubic Upscaling & Baseline &  &  & 28.26 & 0.8277 & - & - & - & - \\
			\hline
		\end{tabular}
	}
	\vspace{2.6mm}
	\caption{\small{Mobile AI 2026 Floating-Point 4K Image Super-Resolution challenge results and final rankings. During the runtime measurements, the models were performing image upscaling from 720$\times$1280 to 2160$\times$3840 pixels on the Dimensity 9500 SoC. Team \textit{ZTE\_cv} is the challenge winner. $^*$ -- At least one model layer falls back to GPU during the inference.}}
	\label{tab:results_float}
\end{table*}

From above 100 registered participants, 15 teams entered the final phase and submitted their results, TFLite models, codes, executables, and factsheets. Tables~\ref{tab:results} and~\ref{tab:results_float} summarize the final challenge results for both tracks and report PSNR, SSIM and runtime numbers for the valid solutions on the final test dataset and on the target evaluation platform. The proposed methods are described in section~\ref{sec:solutions}, and the team members and affiliations are listed in Appendix~\ref{sec:apd:team}.

\subsection{Results and Discussion}

Most of the solutions proposed in this challenge demonstrated an extremely high efficiency, being able to reconstruct 4K image on the target Snapdragon and Dimensity NPUs under 110 ms, significantly surpassing the basic bicubic image upsampling baseline in terms of the resulting image quality. It should also be highlighted that almost all teams performed INT8 model quantization efficiently, and the accuracy drop caused by FP32 $\to$ INT8 conversion is less than 0.2 dB in almost all measured cases.

The winner of the quantized challenge track~--- team \textit{AntYSP}, was able to achieve the best overall results and great network adaptation to the target hardware. This model was able to demonstrate over 230 FPS on the target Qualcomm NPU, which is sufficient even for 4K video super-resolution. Layer re-parametrization, skip connection and quantized-aware training (QAT) were the key factors that allowed to achieve this model high fidelity results while using a shallow architecture. Just slightly lower performance was demonstrated by model from team \textit{Z6}, which was able to process over 180 frames per second on the Snapdragon NPU. The authors also employed layer re-parametrization, skip connections and quantized-aware training (QAT) as well as weight clipping during training to achieve good visual reconstruction image quality.

In the floating-point challenge track, the winner is team \textit{ZTE\_cv}, which solution reached a PSNR score of 30.9 dB while requiring only 110 ms on MediaTek's NPU per image upsampling. Model developed by team \textit{AntYSP} demonstrated a slightly lower PSNR score of 30.7 db while showing the same runtime. Other solutions submitted in this track contained some layers not compatible with the Dimensity NPU, causing them to fall back to GPU-based acceleration, which dramatically affected the resulting runtime values. This also shows the importance of careful model design, ensuring that all model blocks are well optimized and can be successfully executed on the target NPU. 

A notable progress that should be highlighted is the ease of model conversion to the target TFLite format. Tables~\ref{tab:results} and~\ref{tab:results_float} show that many solutions in this challenge were developed using the PyTorch ML framework. Due to the recently presented \textit{litert-torch} plugin~\footnote{\url{https://github.com/google-ai-edge/litert-torch}} developed by the LiteRT team, one can now directly convert the majority of PyTorch models to the TFLite format without the need for TensorFlow / Keras model re-implementation or use of ONNX as an intermediary format. This significantly simplifies the development process of mobile-friendly deep learning architectures for the majority of researchers using PyTorch as their main framework for AI model development and training.

\section{Challenge Methods}
\label{sec:solutions}

\noindent This section describes solutions submitted by all teams participating in the final stage of the MAI 2026 Image Super-Resolution challenge.

\subsection{AntYSP (Tracks 1 and 2)}

Team AntYSP developed two distinct network architectures tailored to the specific requirements of each challenge track.

\begin{figure}[h!]
\centering
\setlength{\tabcolsep}{1pt}
\resizebox{0.8\linewidth}{!}
{
\includegraphics[width=1.0\linewidth]{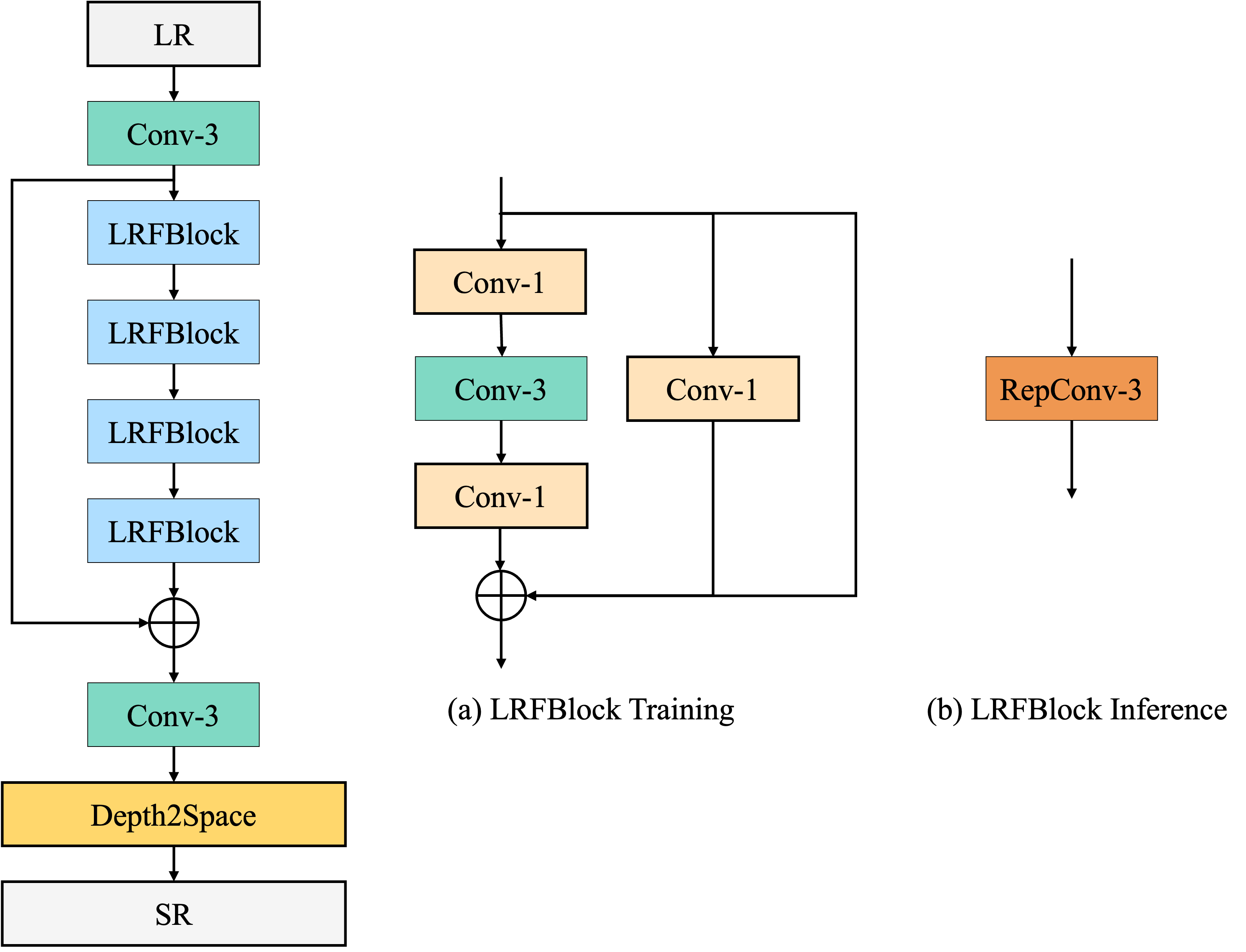}
}
\vspace{0.2cm}
\caption{LiteRF architecture proposed by team AntYSP}
\label{fig:AntYSP}
\end{figure}

\textbf{Track 1:} For the first track, the team proposed the LiteRF architecture, illustrated in Fig.~\ref{fig:AntYSP}. The network starts with a $3\times3$ convolution that expands the input low-resolution image channels to 32. The main feature refinement backbone comprises 4 LiteRFBlock modules, which employ a dual-phase convolution scheme. During training, these blocks utilize a multi-branch topology for enhanced representational capacity, which is then mathematically collapsed into a single efficient $3\times3$ convolution during deployment. Residual bridging connections are included within the pathways to preserve high-frequency information and counteract quantization-induced feature degradation. Following the LiteRFBlock modules, a final $3\times3$ convolution compresses the representations to 27 channels. A dynamic $Min(x, 255)$ clipping operation is applied to prevent activation overflow before a final custom Depth-to-Space operation, specifically aligned with TensorFlow's spatial packing semantics rather than PyTorch's native PixelShuffle, performs the upsampling.

Training was conducted solely on the DIV2K dataset across three stages using the Mean Absolute Error ($L_1$) loss function and the Adam optimizer. Initially, the model was trained from scratch for 800 epochs with a learning rate of $0.001$, a batch size of 96, and low-resolution input patches of size $384\times384$. The second stage involved fine-tuning for 100 epochs with the learning rate decreased to $0.0001$, a batch size of 64, and larger $512\times512$ patches. In the final stage, Quantization-Aware Training (QAT) was performed by inserting FakeQuant nodes. During this phase, the learning rate was further decreased to $0.00001$, while the batch size was explicitly reduced to 16 and the patch size adjusted to $480\times480$ to minimize quantization-induced PSNR loss.

\begin{figure}[h!]
\centering
\setlength{\tabcolsep}{1pt}
\resizebox{0.96\linewidth}{!}
{
\includegraphics[width=1.0\linewidth]{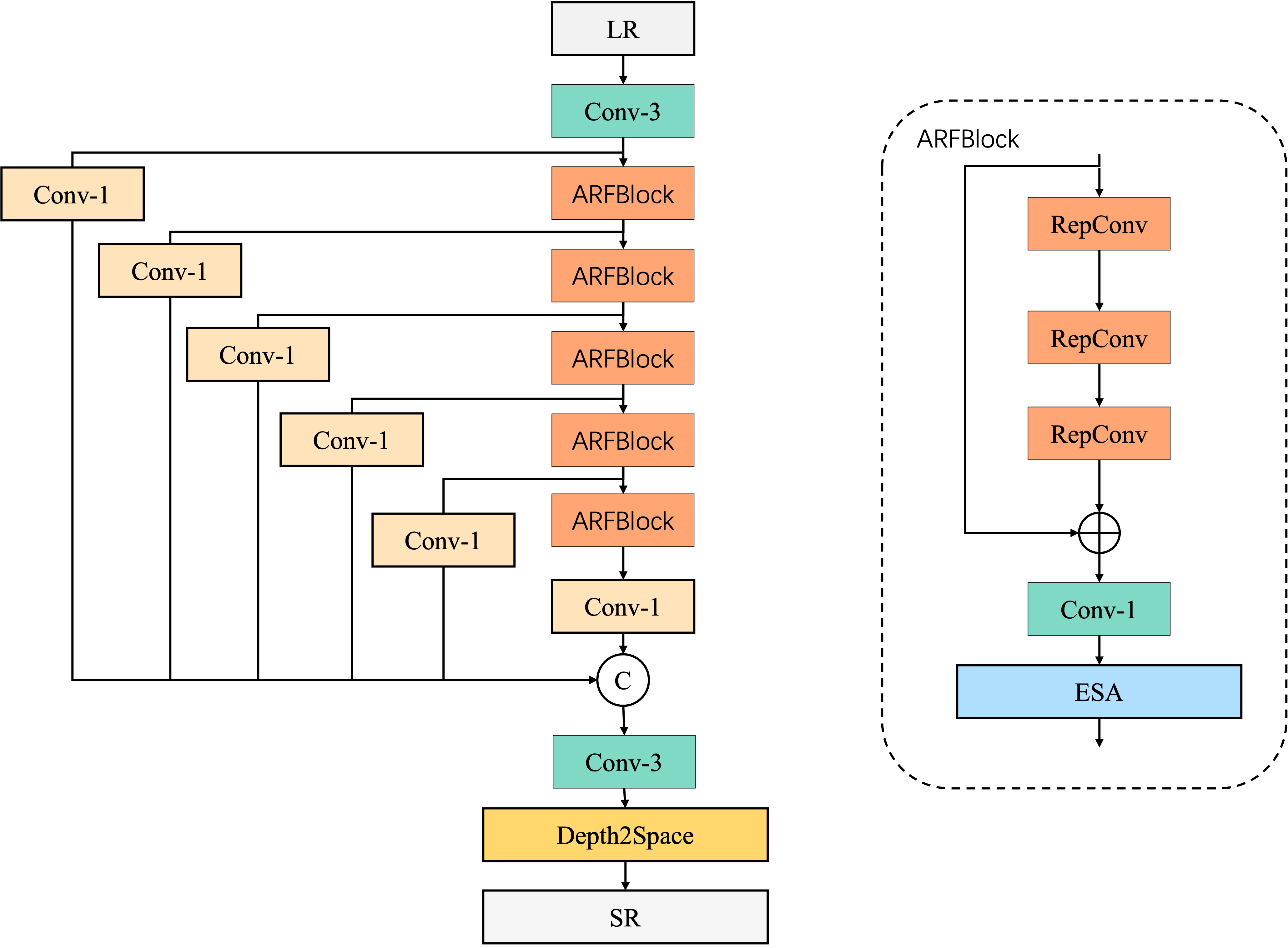}
}
\vspace{0.2cm}
\caption{WYRFBNet architecture proposed by team AntYSP}
\label{fig:AntYSP_track2}
\end{figure}

\textbf{Track 2:} For the second track, the team developed the WYRFBNet architecture, illustrated in Fig.~\ref{fig:AntYSP_track2}. This model begins with a $3\times3$ convolution to extract shallow features from the low-resolution input. The core body of the network consists of multiple cascaded ARFBlock modules. During training, each ARFBlock comprises three sequential $3\times3$ convolutions with ReLU activations, followed by a $1\times1$ projection and an enhanced spatial attention (ESA) module. For efficient deployment, these blocks utilize reparameterization and are merged into a single $3\times3$ convolutions. To facilitate multi-level feature aggregation, the network employs a residual skip mechanism where a $1\times1$ convolution reduces the feature dimension by half to store intermediate representations for each block. After the final ARFBlock, an additional $3\times3$ convolution is applied. All intermediate skip features and the processed backbone output are then concatenated along the channel dimension and fused via a $1\times1$ convolution. Finally, a $3\times3$ convolution projects the features to the required number of channels before a custom-defined Depth-to-Space operation, which replaces PyTorch's native PixelShuffle, performs the upsampling.

Training was conducted across three stages using the Mean Absolute Error ($L_1$) loss function and the Adam optimizer. In the first stage, the model was trained from scratch for 800 epochs with a learning rate of $0.0005$. The second stage involved retraining the model for another 800 epochs, maintaining the same learning rate of $0.0005$. In the final stage, Quantization-Aware Training (QAT) was performed by manually inserting FakeQuant nodes, and the model was fine-tuned with the learning rate decreased to $0.0001$.

\subsection{ZTE\_cv (Track 2)}

\begin{figure}[h!]
\centering
\setlength{\tabcolsep}{1pt}
\resizebox{0.96\linewidth}{!}
{
\includegraphics[width=1.0\linewidth]{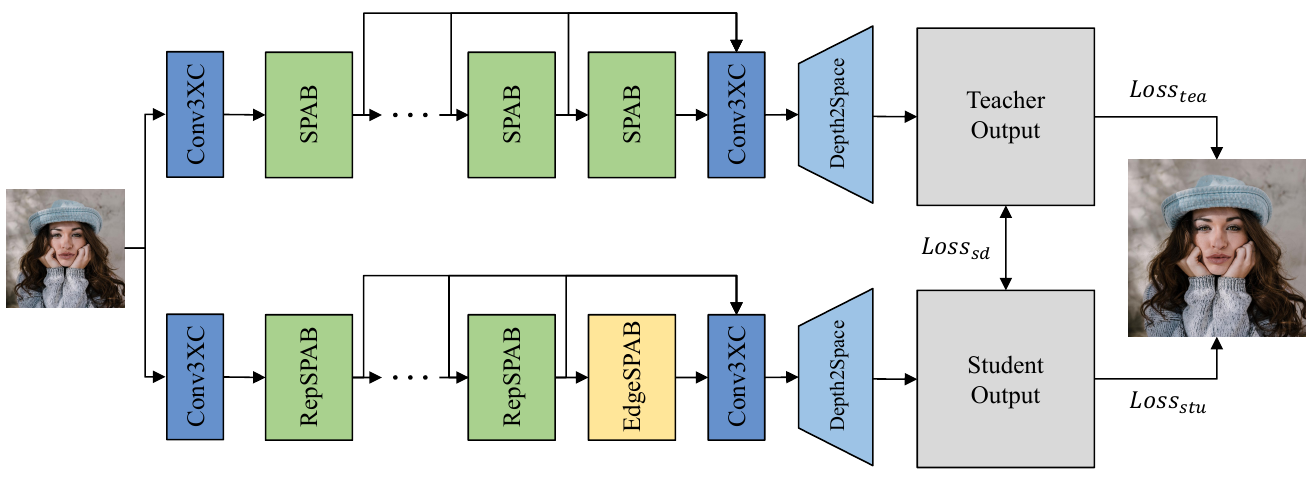}
}
\vspace{0.2cm}
\caption{SPAN hybrid architecture proposed by team ZTE\_cv}
\label{fig:ZTE_cv}
\end{figure}

The SPAN hybrid architecture proposed by team ZTE\_cv, illustrated in Fig.~\ref{fig:ZTE_cv}, is a hierarchical heterogeneous re-parameterization network. The model implements a three-level feature extraction pipeline. Shallow feature extraction is performed by Conv3XC modules. Intermediate feature refinement is handled by RepBlockV6 units, which utilize multi-branch residual re-parameterization. Deep high-frequency enhancement is achieved through EdgeRepBlock modules, which explicitly incorporate edge, gradient, and local geometric priors using specialized branches such as Sobel and Laplacian filters. During inference, all multi-branch structures across these blocks are folded into mathematically equivalent single-path convolutions to maintain deployment efficiency. Multi-layer outputs from these hierarchical stages are then concatenated and compressed via learnable projection layers. Finally, a self-implemented \texttt{pt\_depth2space} operator performs lightweight upsampling by rearranging channel-dimension features into spatial dimensions.

Training was conducted using a three-stage progressive optimization paradigm on curated datasets. In the first stage, foundational training was performed on the cleaned DIV2K dataset using $L_1$ loss, a learning rate of $1\times10^{-3}$, a batch size of 16, and input patches of size $384\times384$. The second stage introduced a curated in-house dataset for domain adaptation, maintaining $L_1$ loss but reducing the learning rate to $5\times10^{-4}$ and increasing the patch size to $480\times480$. The final high-resolution fine-tuning stage utilized a mixed dataset (DIV2K and in-house) with the patch size further increased to $588\times588$, a batch size of 8, and a learning rate of $2\times10^{-4}$. This final stage optimized a joint $L_1$, $L_2$, and self-distillation loss, leveraging an isomorphic 64-channel teacher network to guide the student model through output-level distillation.

\subsection{Z6 (Track 1)}

\begin{figure}[h!]
\centering
\setlength{\tabcolsep}{1pt}
\resizebox{0.96\linewidth}{!}
{
\includegraphics[width=1.0\linewidth]{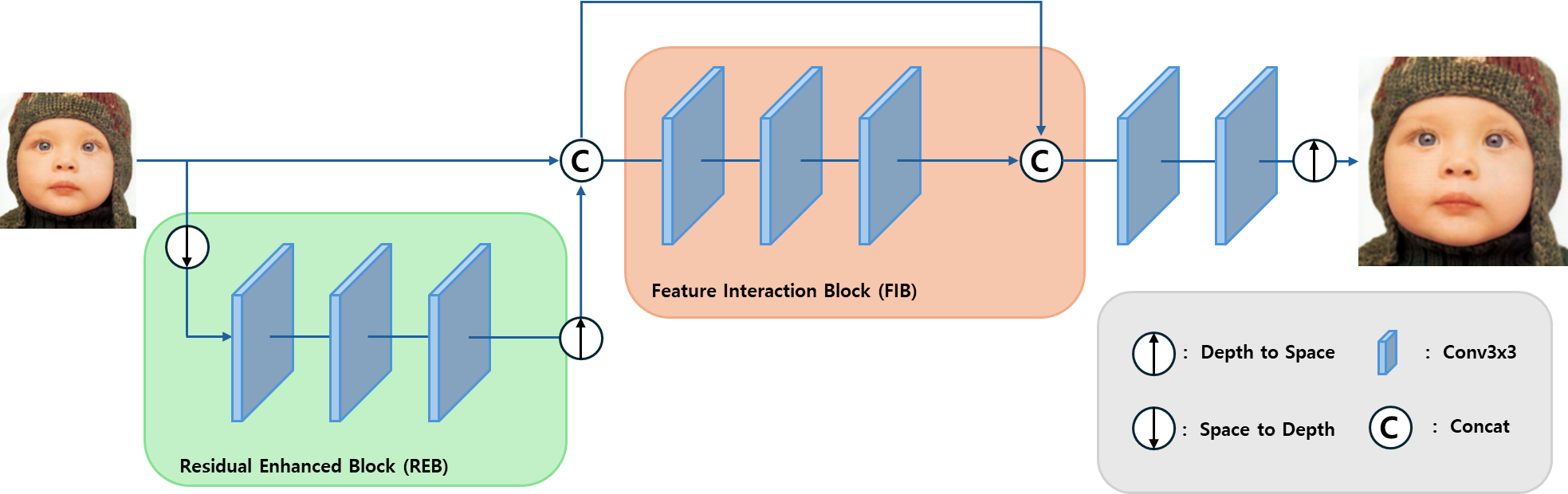}
}
\vspace{0.2cm}
\caption{CNN architecture proposed by Z6 team}
\label{fig:z6}
\end{figure}

Team Z6 introduced the Recursive Concatenation and Enhancement Network (RCENet), depicted in Fig.~\ref{fig:z6}. The architecture routes the input through a cascade of two primary modules: the Residual Enhanced Block (REB) and the Feature Interaction Block (FIB). The foundational components of the network are 32-channel Reparameterized Blocks~\cite{yoon2024casr, gankhuyag2022skip}, which are structurally designed to collapse into single, highly efficient $3\times3$ convolutions during the inference stage. Within the REB, the feature processing flow consists of a Space-to-Depth operation, followed by three $3\times3$ convolutional layers. It concludes with a Depth-to-Space operation alongside a residual connection. These refined features are subsequently fed into the FIB, which integrates three additional $3\times3$ convolutions and leverages a concatenated skip connection at its output. Following these main modules, the pipeline applies two consecutive $3\times3$ convolutions before a final Depth-to-Space layer reconstructs the super-resolved output image.  

To address performance degradation caused by INT8 quantization, the team incorporated specific optimization strategies. They applied weight clipping during training to counteract the highly asymmetric weight distribution, which otherwise exacerbates symmetric quantization errors in TFLite. Furthermore, to prevent inference speed loss, the output values were explicitly clipped just before the final Depth-to-Space layer.  

The model was optimized exclusively on the DIV2K dataset via a three-phase training regimen. The initial phase involved training from scratch for 800 epochs with $L_1$ loss, utilizing an Adam optimizer with a learning rate of $1\text{e-}3$, a batch size of 16, and $128\times128$ input patches, moderated by a cosine warm-up scheduler set to a 0.1\% ratio. In the second phase, the network was fine-tuned for 200 epochs starting from the stage one weights, transitioning to $L_2$ loss to gain extra PSNR and using larger $196\times196$ patches. This stage used a starting learning rate of $2\text{e-}5$, which was halved every 60 epochs via a step scheduler. Finally, Quantization-Aware Training (QAT) was executed for 300 epochs. Initialized from the fine-tuned weights, this phase relied on a DCT-domain $L_1$ loss~\cite{gankhuyag2022skip} to better preserve details under 8-bit precision, maintaining the patch size at $196\times196$ and using a learning rate of $1\text{e-}4$.

\subsection{koreatechV2 (Track 1)}

\begin{figure}[h!]
\centering
\setlength{\tabcolsep}{1pt}
\resizebox{0.8\linewidth}{!}
{
\includegraphics[width=1.0\linewidth]{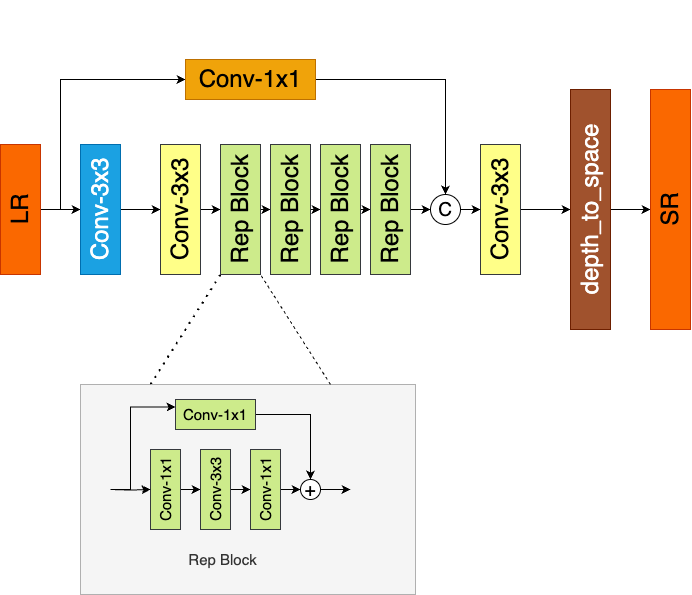}
}
\vspace{0.2cm}
\caption{CCSRN architecture proposed by team koreatechV2}
\label{fig:koreatechV2}
\end{figure}

The architecture proposed by team koreatechV2, illustrated in Fig.~\ref{fig:koreatechV2}, is a streamlined variant of the SCSRN model~\cite{gankhuyag2022skip}. It begins by passing the input low-resolution image through a $3\times3$ convolution. The backbone consists of 6 layers with 32 channels, which includes an initial $3\times3$ convolution followed by four RepConv blocks. During training, each RepConv block employs a sequence of $1\times1$, $3\times3$, and $1\times1$ convolutions in the main path, summed with a parallel $1\times1$ convolution skip path. This design supports increased representational complexity via reparameterization~\cite{bhardwaj2022collapsible}, and is subsequently merged into a single $3\times3$ convolution for lightweight inference. A concatenated skip connection joins the output of the first $3\times3$ convolution (after being processed by a $1\times1$ convolution) with the output of the final RepConv block. A transition layer utilizing a $3\times3$ convolution reduces the feature maps to 27 channels, before a depth-to-space layer performs the upsampling. ReLU activation is applied across all convolutional blocks. Notably, residual paths from the baseline SCSRN~\cite{gankhuyag2022skip} were eliminated to reduce computational overhead.

Training was conducted exclusively on the DIV2K dataset using the Adam optimizer. The optimization strategy utilized Quantization-Aware Training (QAT) fine-tuning to prepare the model for efficient on-device deployment. During training, random cropped patches were extracted from normalized low-resolution images. Data augmentation techniques, including random horizontal and vertical flips, 90-degree rotations, and RGB channel shuffling, were applied to improve model generalization.

\subsection{AIO\_MAI (Track 1)}

\begin{figure}[h!]
\centering
\setlength{\tabcolsep}{1pt}
\resizebox{0.7\linewidth}{!}
{
\includegraphics[width=0.8\linewidth]{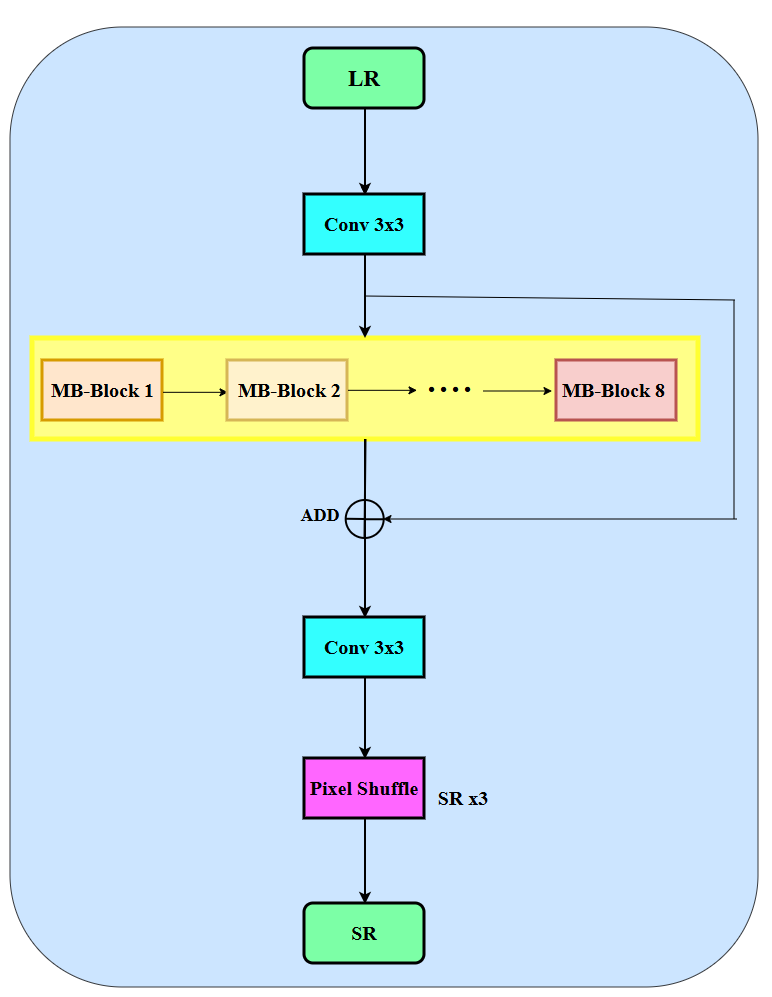}
}
\vspace{0.2cm}
\caption{Architecture proposed by team AIO\_MAI}
\label{fig:aiomai}
\end{figure}

The architecture proposed by team AIO\_MAI, illustrated in Fig.~\ref{fig:aiomai}, follows an extract-refine-upsample paradigm specifically designed for efficient end-to-end TensorFlow Lite INT8 deployment. The network initiates processing with a lightweight convolution that maps the RGB input into a feature space. The core backbone consists of stacked multi-branch blocks inspired by MobileOne~\cite{mobileone2022}. During training, each of these blocks employs a highly expressive multi-branch topology consisting of four parallel $3\times3$ convolutional branches, one $1\times1$ branch, and an identity branch, which are all summed before passing through a ReLU activation. To eliminate runtime overhead, structural reparameterization is utilized at deployment, folding the complex multi-branch structure into a single equivalent $3\times3$ convolution. A global skip connection at the feature level allows the backbone to focus on high-frequency refinement by directly passing low-frequency content. Processing is strictly maintained in the low-resolution space to minimize FLOPs, with upsampling deferred to the final step where features are projected to 27 channels and rearranged using a PixelShuffle operation. Furthermore, dynamic weight clipping is applied to suppress activation outliers and improve robustness prior to quantization.

The model was trained over three curriculum stages utilizing the Adam optimizer. Stage one focused on base reconstruction, training the model for 600 epochs using $L_1$ loss, a learning rate of $10^{-3}$ with a cosine warmup schedule, and low-resolution patches of $128\times128$. Stage two fine-tuned the model for 200 epochs using a larger $160\times160$ patch size and a learning rate of $3.0\times10^{-5}$. This fidelity-oriented stage employed a Charbonnier loss, a DCT frequency-domain loss, and an output-level Knowledge Distillation (KD) loss guided by a pre-trained MambaIRv2Light~\cite{guo2024mambairv2} teacher model. Before proceeding to the final stage, the best checkpoint underwent Batch Normalization recalibration over 64 mini-batches to ensure accurate running statistics, followed by branch fusion. Stage three executed Quantization-Aware Training (QAT) for 150 epochs directly on the reparameterized deploy graph using PyTorch FX graph mode with a QNNPACK backend. During this phase, mixed precision was disabled to maintain fake-quantization stability, the learning rate was reduced to $2.0\times10^{-6}$, the patch size was adjusted to $144\times144$, and a batch size of 1 was used.

\subsection{IN2GM (Track 1)}

\begin{figure}[h!]
\centering
\setlength{\tabcolsep}{1pt}
\resizebox{0.96\linewidth}{!}
{
\includegraphics[width=1.0\linewidth]{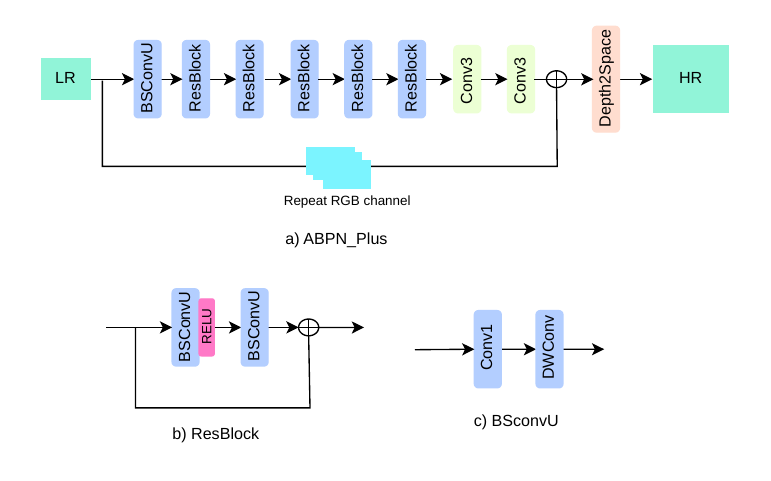}
}
\vspace{0.2cm}
\caption{ABPN Plus architecture proposed by team IN2GM}
\label{fig:IN2GM}
\end{figure}

The ABPN Plus architecture proposed by team IN2GM, illustrated in Fig.~\ref{fig:IN2GM}, builds upon the Anchor-Based Plain Network framework~\cite{du2021anchor}. The input low-resolution image is processed through two parallel paths. In the bottom path, the input RGB channels are simply replicated to provide a baseline estimate. The top main path begins with a Blueprint Separable Convolution (BSConvU) layer~\cite{li2022blueprint}, which decomposes standard convolution into a $1\times1$ convolution followed by a depthwise convolution to improve computational efficiency. This initial extraction is followed by a sequence of five stacked residual blocks. Each residual block contains two BSConvU layers separated by a ReLU activation, along with a skip connection. After the residual blocks, two consecutive convolutions are applied to refine the features. This predicted residual from the top path is then added to the replicated input baseline from the bottom path, and a Depth-to-Space operation performs the final upsampling to generate the high-resolution image.

Training was conducted using the DIV2K dataset and the $L_1$ loss function. The model was optimized for 600 epochs using the Adam optimizer with an initial learning rate of $1 \times 10^{-3}$, which was halved every 120 epochs. Randomly extracted patches of size $64 \times 64$ and a batch size of 16 were employed throughout this phase. To optimize for mobile deployment, a subsequent Quantization-Aware Training (QAT) stage was conducted for 200 epochs. During QAT, the initial learning rate was set to $1 \times 10^{-3}$ and halved every 50 epochs to help the model adapt its weights to low-precision computations for inference.

\section{Additional Literature}

An overview of the past challenges on mobile-related tasks together with the proposed solutions can be found in the following papers:

\begin{itemize}
\item Image Super-Resolution:\, \cite{ignatov2018pirm,cai2019ntire,timofte2018ntire,lugmayr2020ntire}
\item Learned End-to-End ISP:\, \cite{ignatov2019aim,ignatov2020aim,wirzberger2022lan,ignatov2022microisp,ignatov2022pynetv2}
\item RGB Image Enhancement:\, \cite{ignatov2018pirm,ignatov2019ntire,ignatov2017dslr,ignatov2018wespe}
\item Bokeh Effect Rendering:\, \cite{ignatov2019aimBokeh,ignatov2020aimBokeh,ignatov2020rendering}
\item Image Denoising:\, \cite{abdelhamed2019ntire,abdelhamed2020ntire,flepp2024real}
\end{itemize}

\section*{Acknowledgements}

We thank AI Witchlabs, ETH Z\"urich, and the University of W\"urzburg (Computer Vision Lab), the organizers and sponsors of this Mobile AI 2026 challenge.

\appendix
\section{Teams and Affiliations}
\label{sec:apd:team}

\bigskip

\subsection*{Mobile AI 2026 Team}
\noindent\textit{\textbf{Title: }}\\ Mobile AI 2026 4K Image Super-Resolution Challenge\\
\noindent\textit{\textbf{Members:}}\\ Andrey Ignatov$^{1,3}$ \textit{(andrey@vision.ee.ethz.ch)}, Georgy Perevozchikov$^2$ \textit{(georgii.perevozchikov@uni-wuerzburg. de)}, Radu Timofte$^{2,3}$ \textit{(radu.timofte@uni-wuerzburg.de)}\\
\noindent\textit{\textbf{Affiliations: }}\\
$^1$ Computer Vision Lab, ETH Z\"urich, Switzerland\\
$^2$ Computer Vision Lab, University of W\"urzburg, Germany\\
$^3$ AI Witchlabs, Switzerland\\

\subsection*{AntYSP}
\noindent\textit{\textbf{Members:}}\\ Fubin Ren$^1$ \textit{(wangqiu.rfb@gmail.com)}, Shihao Wang$^1$\\
\noindent\textit{\textbf{Affiliations: }}\\
$^1$ Ant Group, China\\

\subsection*{ZTE\_cv}
\noindent\textit{\textbf{Members:}}\\
Liwen Zhang$^1$ \textit{(zhang.liwen3@zte.com.cn)}, Zhe Xu$^1$ \textit{(xu.zhe3@zte.com.cn)}, Zhipei Lei$^1$ \textit{(lei.zhipei@zte.com.cn)}, Minjian Zhang$^1$ \textit{(zhang.minjian2@zte.com.cn)}, Kailing Tang$^1$ \textit{(tang.kailing@zte.com.cn)}, Dingyong Gou$^1$ \textit{(gou.dingyong@zte.com.cn)}, Yanlin Wu$^1$ \textit{(wu.yanlin@zte.com.cn)}, Cong Li$^1$ \textit{(li.cong4@zte.com.cn)}, Xiaohui Cui$^1$ \textit{(cui.xiaohuil@zte.com.cn)}\\
\noindent\textit{\textbf{Affiliations: }}\\
$^1$ ZTE Corporation, China

\subsection*{Z6}
\noindent\textit{\textbf{Members:}}\\ Kihwan Yoon$^1$ \textit{(rlghksdbs@gmail.com)}, Ganzorig Gankhuyag$^2$\\
\noindent\textit{\textbf{Affiliations: }}\\
$^1$ BLUEDOT, Korea\\
$^2$ Korea Electronics Technology Institutes (KETI), Korea\\

\subsection*{koreatech V2}
\noindent\textit{\textbf{Members:}}\\ Heo MyeongCheol$^1$ \textit{(soon3771@koreatech.ac.kr)}, Yonghyun Park$^1$, Juho Jung$^1$, Sungjei Kim$^1$\\
\noindent\textit{\textbf{Affiliations: }}\\
$^1$ Korea University of Technology \& Education (KOREATECH), South Korea\\

\subsection*{AIO MAI}
\noindent\textit{\textbf{Members:}}\\ Phuong-Nam Nguyen-Pham$^1$ \textit{(phuongnamdn05@gmail.com)}, Nam-Tien Le$^1$, Tinh-Anh Nguyen-Nhu$^1$\\
\noindent\textit{\textbf{Affiliations: }}\\
$^1$ AIVN, Vietnam\\

\subsection*{IN2GM}
\noindent\textit{\textbf{Members:}}\\ Matin Fazel$^1$ \textit{(matin.fazel@concordia.ca)}, Abdelhak Bentaleb$^1$ \textit{(abdelhak.bentaleb@concordia.ca)}\\
\noindent\textit{\textbf{Affiliations: }}\\
$^1$ Concordia University, Canada\\

\subsection*{LoongSR}
\noindent\textit{\textbf{Members:}}\\ Xinrui Li$^1$ \textit{(felix.leeovo7@gmail.com)}, Xusheng He$^1$, Jianlong Wu$^1$\\
\noindent\textit{\textbf{Affiliations: }}\\
$^1$ Harbin Institute of Technology (Shenzhen), China\\

\subsection*{Godzilla}
\noindent\textit{\textbf{Members:}}\\ Mohsin Ikram$^1$ \textit{(mohsin271998@gmail.com)}, Dmitry Ignatov$^1$\\
\noindent\textit{\textbf{Affiliations: }}\\
$^1$ Computer Vision Lab, CAIDAS \& IFI, University of Würzburg, Germany\\

\subsection*{Sunrise2046}
\noindent\textit{\textbf{Members:}}\\ Hao Chen$^1$ \textit{(2658835314@qq.com)}, Long Sun$^1$, Jiangxin Dong$^1$, Jinshan Pan$^1$, Jinhui Tang$^1$\\
\noindent\textit{\textbf{Affiliations: }}\\
$^1$ Nanjing University of Science and Technology, China\\

\subsection*{BVIISR}
\noindent\textit{\textbf{Members:}}\\ Siyue Teng$^1$ \textit{(eq21135@bristol.ac.uk)}, Yuxuan Jiang$^1$, Fan Zhang$^1$, David Bull$^1$
\noindent\textit{\textbf{Affiliations: }}\\
$^1$ Visual Information Laboratory, University of Bristol, UK\\

\subsection*{Sony-PolyU-LUT}
\noindent\textit{\textbf{Members:}}\\
Zhe Xiao$^{1,2}$ \textit{(xiao-zhe.xiao@connect.polyu.hk)}, Jun Xiao$^2$, Zhi-Song Liu$^3$, Jiayi Pei$^1$, Songtao Wu$^1$, Hong Shang$^1$, Kuanhong Xu$^1$\\
\noindent\textit{\textbf{Affiliations: }}\\
$^1$ Sony R\&D China, China\\
$^2$ The Hong Kong Polytechnic University, Hong Kong\\
$^3$ Lappeenranta-Lahti University of Technology, Finland\\

{\small
\bibliographystyle{ieeenat_fullname}
\bibliography{egbib}
}

\end{document}